\pdfoutput=1

\documentclass[11pt]{article}

\usepackage[final]{acl}

\usepackage{times}
\usepackage{latexsym}
\usepackage{booktabs}
\usepackage{amsmath}
\usepackage{color}
\usepackage{colortbl}
\usepackage{mathtools}
\usepackage{enumitem}

\usepackage[T1]{fontenc}

\usepackage[utf8]{inputenc}

\usepackage{microtype}

\usepackage{inconsolata}

\usepackage{graphicx}

\usepackage{tikz}
\usetikzlibrary{positioning, shapes.geometric, arrows.meta, decorations.pathreplacing, calc, plotmarks}

\usepackage{pgfplots}
\usepackage{amsmath}
\usepackage{booktabs}
\usepackage{hyperref}
\usepackage{caption}

\pgfplotsset{compat=newest}

\newcommand{\first}[1]{\cellcolor{green!40}{#1}}
\newcommand{\second}[1]{\cellcolor{yellow!40}{#1}}
\newcommand{\third}[1]{\cellcolor{orange!40}{#1}}
\title{STAGE: Simplified Text-Attributed Graph Embeddings Using Pre-trained LLMs}

\author{
 \textbf{Aaron Zolnai-Lucas\textsuperscript{1}\thanks{Authors contributed equally.}},
 \textbf{Jack Boylan\textsuperscript{1}\footnotemark[1]},
 \textbf{Chris Hokamp\textsuperscript{1}},
 \textbf{Parsa Ghaffari\textsuperscript{1}}
\\
\\
 \textsuperscript{1}Quantexa,
\\
 \small{
   \textbf{Correspondence:} {\texttt{\{firstname\}\{lastname\}@quantexa.com}}
 }
}

\begin{document}
\maketitle
\begin{abstract}
We present Simplified Text-Attributed Graph Embeddings (STAGE), a straightforward yet effective method for enhancing node features in Graph Neural Network (GNN) models that encode Text-Attributed Graphs (TAGs). Our approach leverages Large-Language Models (LLMs) to generate embeddings for textual attributes. STAGE achieves competitive results on various node classification benchmarks while also maintaining a simplicity in implementation relative to current state-of-the-art (SoTA) techniques. We show that utilizing pre-trained LLMs as embedding generators provides robust features for ensemble GNN training, enabling pipelines that are simpler than current SoTA approaches which require multiple expensive training and prompting stages. We also implement diffusion-pattern GNNs in an effort to make this pipeline scalable to graphs beyond academic benchmarks.
\end{abstract}

\section{Introduction}
\label{sec:introduction}

A Knowledge Graph (KG) typically includes entities (represented as nodes), relationships between entities (represented as edges), and attributes of both entities and relationships \cite{Ehrlinger2016TowardsAD}. These attributes, referred to as metadata, are often governed by a domain-specific ontology, which provides a formal framework for defining the types of entities and relationships as well as their properties. KGs can be used to represent structured information about the world in diverse settings, including medical domain models \cite{Kon2023SNOMEDCA}, words and lexical semantics \cite{10.1145/219717.219748wordnet}, and commercial products \cite{Chiang_2019_ogbn_products}.

Text-Attributed Graphs (TAGs) can be viewed as a subset of KGs, where some node and edge metadata is represented by unstructured or semi-structured natural language text \cite{yang2023graphformers}. Examples of unstructured data values in TAGs could include the research article text representing the nodes of a citation graph, or the content of social media posts that are the nodes of an interaction graph extracted from a social media platform. Many real-world datasets are naturally represented as TAGs, and studying how to best represent and learn using these datasets has received attention from the fields of graph learning, natural language processing (NLP), and information retrieval.

\paragraph{Graph Learning and LLMs}
With the emergence of LLMs as powerful general purpose reasoning agents, there has been increasing interest in integrating KGs with LLMs \cite{Pan_2024}. 
Current SoTA approaches combining graph learning with (L)LMs follow either an \textbf{iterative} or a \textbf{cascading} method. \textit{Iterative} methods involve jointly training an LM and a GNN for the given task. While this approach can produce a task-specific feature space, it may be complex and resource-intensive, particularly for large graphs. In contrast, \textit{cascading} methods first apply an LM to extract node features which are then used by a downstream GNN model. Cascading models demonstrate excellent performance on TAG tasks \cite{he2024harnessing, duan2023simteg}, although they often require multiple stages of training targeted at each pipeline component. More recent cascading techniques implement an additional step, known as text-level enhancement \cite{chen2024exploring}, whereby textual features are augmented using an LLM.


\begin{figure*}[htbp]
  \centering
  \resizebox{\textwidth}{!}{%
    \includegraphics{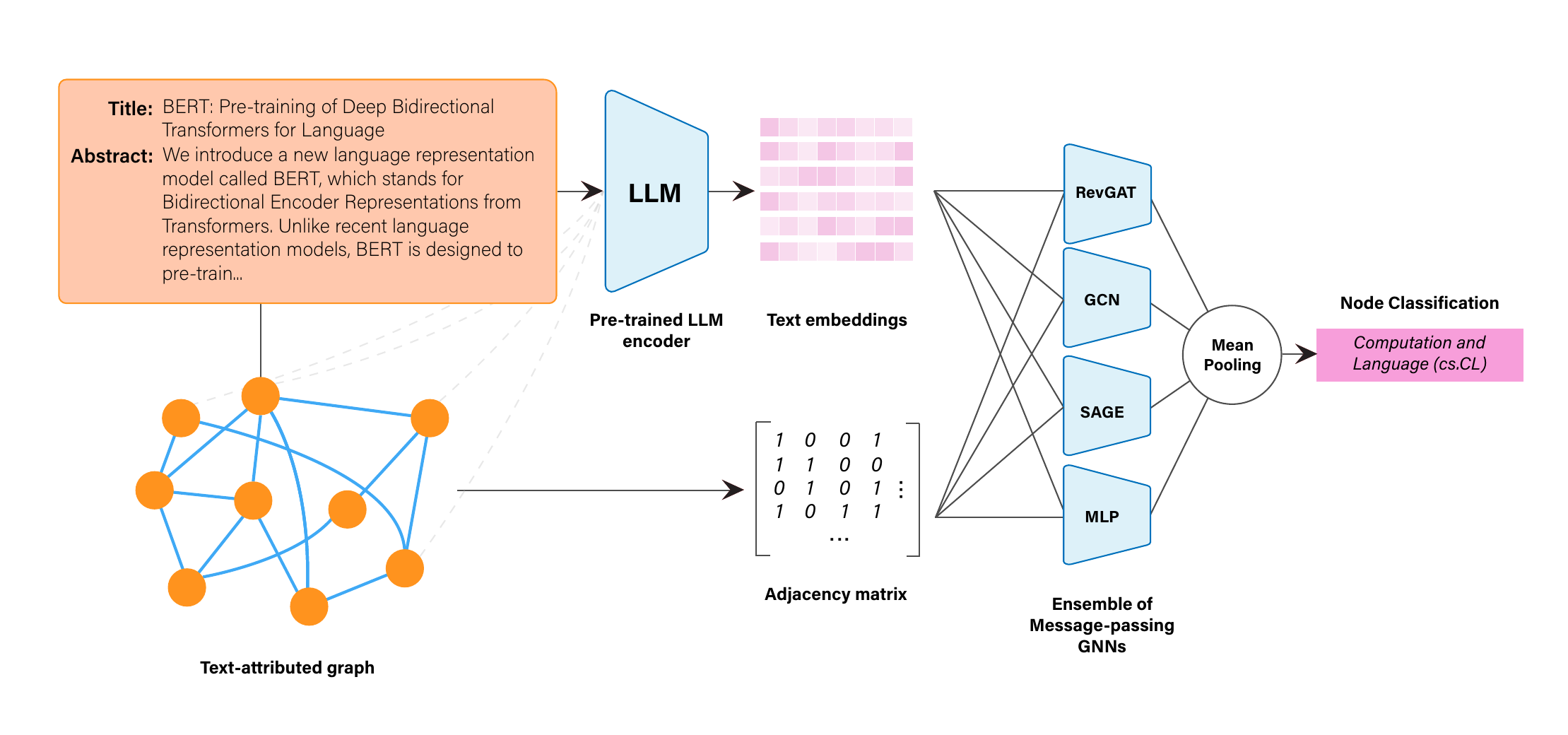}
  }
  \caption{Our proposed approach to node classification. Firstly, the textual attributes of the input graph nodes are encoded using an off-the-shelf LLM. The text embeddings will be used alongside the graph adjacency matrix as input to train a downstream ensemble of GNNs. GNN predictions are then mean-pooled to obtain the final prediction.}
  \label{fig:main-diagram}
\end{figure*}

\paragraph{Simplifying Node Representation Generation}

To the best of our knowledge, all existing cascading approaches require multiple rounds of data generation or finetuning to achieve satisfactory results on TAG tasks \cite{he2024harnessing, duan2023simteg, chen2024exploring}. This bottleneck increases the difficulty of applying such methods to real-world graphs. Our proposed method, STAGE, aims to simplify existing approaches by foregoing LM finetuning, and only making use of a single pre-trained LLM as the node embedding model, without data augmentation via prompting. We study possible configurations of this simplified pipeline and demonstrate that this method achieves competitive performance while significantly reducing the complexity of training and data preparation.

\paragraph{Scalable GNN Architectures}
The exponentially growing receptive field required during training of most message-passing GNNs is another bottleneck in both cascading and iterative approaches, becoming computationally intractable for large graphs \citep{duan2023comprehensive,liu2024scalable}. Because we wish to study approaches that can be applied in real-world settings, we also explore the implementation of diffusion-pattern GNNs, such as Simple-GCN \cite{simpleGCNwu} and SIGN \cite{frasca2020sign}, which may enable STAGE to be applied to much larger graphs beyond the relatively small academic benchmarks. Our code is available at \url{https://github.com/aaronzo/STAGE}.

Concretely, this work studies several ways to make learning on TAGs more efficient and scalable:
\begin{itemize}[itemsep=0.1em]
    \item \textbf{Single Training Stage:} We perform ensemble GNN training with a fixed LLM as the node feature generator, which significantly reduces training time by eliminating the need for multiple large model training runs.
    \item \textbf{No LLM Prompting:} We do not prompt an LLM for text-level augmentations such as predictions or explanations. Instead, we use only the text attributes provided in the dataset.
    \item \textbf{Direct Use of LLM as Text Embedding Model:} Using an off-the-shelf LLM as the embedding model makes this method adaptable to new models and datasets. We study several alternative base models for embedding generation.
    \item \textbf{Diffusion-pattern GNN implementation:} We contribute an investigation into diffusion-pattern GNNs which enable this method to scale to larger graphs.
\end{itemize}

\noindent The rest of the paper is organized as follows: section \ref{sec:background} gives an overview of related work, section \ref{sec:approach} discusses our approach in detail, section \ref{sec:experiments} studies the performance of STAGE in various settings, and section \ref{sec:analysis} is a discussion of the experimental results.

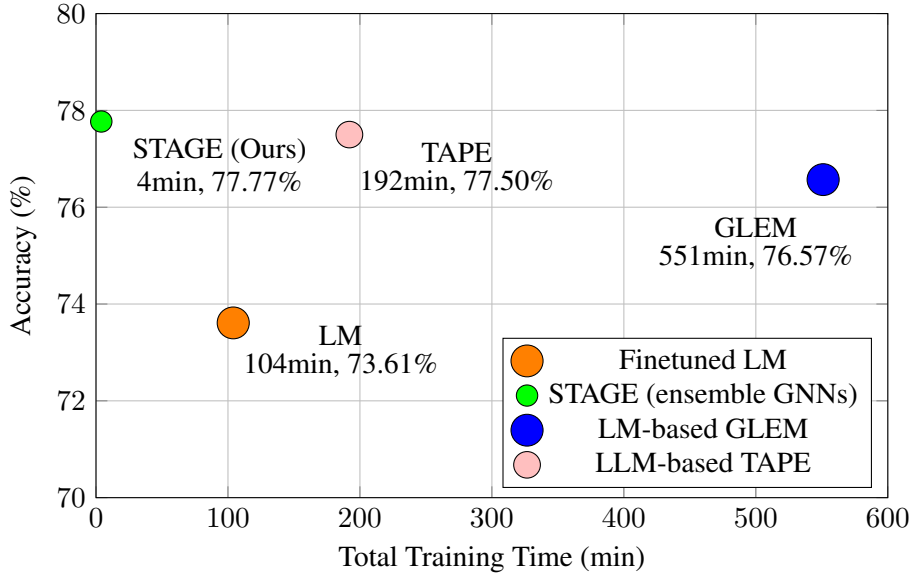
\begin{figure*}[h!]
\centering
\begin{tikzpicture}
    \begin{axis}[
        width=12cm, height=8cm,
        xlabel={Total Training Time (min)},
        ylabel={Accuracy (\%)},
        xmin=0, xmax=600,
        ymin=70, ymax=80,
        grid=both,
        legend style={
            at={(0.98,0.02)}, 
            anchor=south east, 
            draw=black, 
            fill=white, 
            align=left 
        },
        scatter/classes={
            lm={mark=*,draw=black,fill=orange, mark size=6}, 
            stage={mark=*,draw=black,fill=green, mark size=4}, 
            glem={mark=*,draw=black,fill=blue, mark size=6}, 
            tape={mark=*,draw=black,fill=pink, mark size=5} 
        }
    ]

    \addplot[scatter,only marks,scatter src=explicit symbolic]
        coordinates {
            (104, 73.61) [lm]
            (4, 77.77) [stage]
            (551, 76.57) [glem]
            (192, 77.50) [tape]
        };

    \node[anchor=west] at (axis cs: 104, 73.01) {\shortstack{LM\\104min, 73.61\%}};
    \node[anchor=west] at (axis cs: 20, 76.83) {\shortstack{STAGE (Ours)\\4min, 77.77\%}};
    \node[anchor=north] at (axis cs: 500, 76.00) {\shortstack{GLEM\\551min, 76.57\%}};
    \node[anchor=west] at (axis cs: 192, 76.80) {\shortstack{TAPE\\192min, 77.50\%}};


    \legend{Finetuned LM,STAGE (ensemble GNNs),LM-based GLEM,LLM-based TAPE}
    \end{axis}
\end{tikzpicture}
\caption{The performance trade-off between node classification accuracy and total training time on \texttt{ogbn-arxiv} for SoTA LM-GNN methods. The STAGE model uses text embeddings generated from Salesforce-Embedding-Mistral and an ensemble of GNNs (GCN, SAGE and RevGAT) and MLP. The size of each marker indicates the total number of trainable parameters. Figure adapted from \cite{he2024harnessing}.}
\label{fig:tradeoff}
\end{figure*}

\section{Background}
\label{sec:background}


\paragraph{Text-Attributed Graphs}

\citet{yan2023comprehensive} suggest that integrating topological data with textual information can significantly improve the learning outcomes on various graph-related tasks. \citet{chien2022node} incorporate graph structural information into the pre-training stage of pre-trained language models (PLMs), achieving improved performance albeit with additional training overhead, while \citet{liu2023all} further adopt sentence embedding models to unify the text-attribute and graph structure feature space, proposing a unified model for diverse tasks across multiple datasets.



\paragraph{LLMs as Text Encoders}




General purpose text embedding models, used in both finetuned and zero-shot paradigms, are a standard component of modern NLP pipelines \citep{mikolov2013distributed,pennington-etal-2014-glove,reimers-2019-sentence-bert}. As LLMs have emerged as powerful zero-shot agents, many studies have considered generating text embeddings as an auxiliary output \cite{muennighoff2022sgpt,mialon2023augmented}. \citet{behnamghader2024llm2vec} introduce LLM2Vec, an unsupervised method to convert LLMs into powerful text encoders by using bidirectional attention, masked next token prediction and contrastive learning, achieving state-of-the-art performance on various text embedding benchmarks.

\paragraph{Language Models and GNNs}

Graph Neural Networks have been successfully applied to node classification and link prediction tasks, demonstrating improved performance when combined with textual features from nodes \cite{kipf2017semisupervised, li2022house}. Several studies show that finetuning pre-trained Language Models (PLMs), such as BERT \cite{devlin2019bert} and DeBERTa \cite{he2021deberta}, enhances GNN performance by leveraging textual node features \cite{chen2024exploring, duan2023simteg, he2024harnessing}.

Recent research has explored the integration of LLMs with GNNs, particularly for TAGs. LLMs contribute deep semantic understanding and commonsense knowledge, potentially boosting GNNs' effectiveness on downstream tasks. However, combining LLMs with GNNs poses computational challenges. Techniques like GLEM \cite{glem} use the Expectation Maximization framework to alternate updates between LM and GNN modules.

Other approaches include the TAPE method, which uses GPT \cite{openai_chatgpt, openai2024gpt4} models for data augmentation, enhancing GNN performance through enriched textual embeddings \cite{he2024harnessing}. SimTeG demonstrates that parameter-efficient finetuning (PEFT) PLMs can yield competitive results \cite{duan2023simteg}. \cite{ye-etal-2024-language-is-all} suggest that finetuned LLMs can match or exceed state-of-the-art GNN performance on various benchmarks.

Building on these insights, the STAGE method focuses on efficient and scalable learning for TAGs by utilizing zero-shot capabilities of LLMs to generate representations without extensive task-specific tuning or auxiliary data generation.

\section{Approach}
\label{sec:approach}

Our cascading approach consists of two steps: 

\begin{itemize}[itemsep=0.1em]
    \item A zero-shot LLM-based embedding generator is used to encode the title and abstract (or equivalent textual attribute) of each node. We denote the generated node embeddings as $\mathbf{\mathcal{X}}$.
    \item An ensemble of GNN architectures are trained on $\mathbf{\mathcal{X}}$, and their predictions are mean-pooled to obtain the final node predictions.
\end{itemize}

\noindent Ensembling the predictions from multiple GNN architectures was motivated by our observation of strong performance by different models across different datasets.


\subsection{Text Embedding Retrieval}
\label{sec:text-embedding-retrieval}

For the text embedding model, we select a general-purpose embedding LLM that ranks highly on the Massive Text Embedding Benchmark (MTEB) Leaderboard\footnote{\url{https://huggingface.co/spaces/mteb/leaderboard}}. Specifically, we evaluate \texttt{gte-Qwen1.5-7B-instruct}, \texttt{LLM2Vec-Meta-Llama-3-8B-Instruct}, and \texttt{SFR-Embedding-Mistral}. MTEB ranks embedding models based on their performance across a wide variety of information retrieval, classification and clustering tasks. This model is used out-of-the-box without any finetuning. An appealing aspect of LLM-based embeddings is the possibility to add instructions alongside input text to bias the embeddings for a given task. We empirically evaluate the effect of instruction biased embeddings is in Table \ref{tab:instruction-bias-results} of section \ref{sec:experiments}.

Node representations $\mathbf{\mathcal{X}}$ are generated using only the title and abstract, or equivalent textual node attributes, omitting the LLM predictions and explanations provided by \cite{he2024harnessing}. $\mathbf{\mathcal{X}}$ will then be used as enriched node feature vectors for training a downstream GNN ensemble.

\subsection{GNN Training}

Using the previously generated embeddings $\mathbf{\mathcal{X}}$ as node features, we train an ensemble of GNN models on the node classification task:
\begin{equation}
    \text{Loss}_{\text{cls}} = \mathcal{L}_{\theta}\left(\phi(\text{GNN}(\mathcal{X}, \mathcal{A})), \mathbf{Y}\right),
\end{equation}

\noindent where $\phi(\cdot)$ is the classifier, $\mathbf{\mathcal{A}}$ is the adjacency matrix of the graph and $\mathbf{\mathcal{Y}}$ is the label. For the GNN architectures we choose GCN \cite{kipf2017semisupervised}, SAGE \cite{graphsage} and RevGAT \cite{li2022training}. We also evaluate a multi-layer perceptron (MLP) \cite{haykin1994neural} among our GNN models. To combine the predictions from each of the $K$ models in the ensemble, we compute the mean prediction as follows:

\begin{equation}
    \bar{\mathbf{p}} = \frac{1}{K} \sum_{k=1}^{K} \mathbf{p}_k,
\end{equation}

\noindent Cross-entropy loss is used to compute the loss value.


\paragraph{Diffusion-based GNNs}

For a graph $G$ with node features $\mathbf{\mathcal{X}}$, a diffusion operator is a matrix $A_{\text{OP}}$ with the same dimensions as the adjacency matrix of $G$. Diffused features $\mathcal{H}$ are then calculated via $\mathcal{H} = A_{\text{OP}}\mathcal{X}$.

We explored Simple-GCN \cite{simpleGCNwu} and SIGN \cite{frasca2020sign}, both of which employ adjacency-based diffusion operators to pre-aggregate features across the graph before training. SIGN is a generalization of Simple-GCN, to extend to Personalized-PageRank \cite{Page1998PageRank} and triangle-based operators. This allows expensive computation to be carried out by distributed computing clusters or efficient sparse graph routines such as GraphBLAS \cite{GraphBLAS7}, which do not need to back-propagate through graph convolution. The prediction head can then be a shallow MLP or logistic regression. We provide implementation specifics in appendix section \ref{sec:implementing-diffusion-operators} to ensure repeatability.

\subsection{Parameter-efficient Finetuning LLM}
Motivated by the node classification performance gains seen by \cite{duan2023simteg} using PEFT, we finetune an LLM on the node classification task. Concretely, we use an LLM embedding model with a low-rank adapter (LoRA) \cite{hu2021lora} and a densely connected classifier head. The pre-trained LLM weights remain frozen as the model trains on input text $T$ to reduce loss according to:
\begin{equation}
    \text{Loss}_{\text{cls}} = \mathcal{L}(\phi(\text{LLM}(T)), Y)
\end{equation}

\noindent where $\phi(\cdot)$ is the classifier head and $Y$ is the label. Again, we use cross-entropy loss to compute the loss value.

\section{Experiments}
\label{sec:experiments}

\begin{table*}[h!]
\centering
\setlength{\tabcolsep}{4pt}
\small
\begin{tabular}{lcccccccc}
\toprule
\textbf{Dataset} & \textbf{Method} & $h_{\text{shallow}}$ & $h_{\text{GIANT}}$ & GPT3.5 & $LM_{\text{finetune}}$ & $h_{\text{TAPE}}$ & $h_{\text{STAGE}} \textbf{(OURS)}$ \\
\midrule
\textbf{Cora} & MLP & 0.6388 $\pm$ 0.0213 & 0.7196 $\pm$ 0.0000 & 0.6769 & 0.7606 $\pm$ 0.0378 & 0.8778 $\pm$ 0.0485 & 0.7680 $\pm$ 0.0228 \\
& GCN & 0.8911 $\pm$ 0.0015 & 0.8423 $\pm$ 0.0053 & 0.6769 & 0.7606 $\pm$ 0.0378 & \third{0.9119 $\pm$ 0.0158} & 0.8704 $\pm$ 0.0105 \\
& SAGE & 0.8824 $\pm$ 0.0009 & 0.8455 $\pm$ 0.0028 & 0.6769 & 0.7606 $\pm$ 0.0378 & \first{0.9290 $\pm$ 0.0307} & 0.8722 $\pm$ 0.0063 \\
& RevGAT & 0.8911 $\pm$ 0.0000 & 0.8353 $\pm$ 0.0038 & 0.6769 & 0.7606 $\pm$ 0.0378 &  \second{0.9280 $\pm$ 0.0275} & 0.8639 $\pm$ 0.0129 \\
& \textbf{Ensemble} & - & - & - & - & - & 0.8824 $\pm$ 0.0155 \\
\midrule
\textbf{PubMed} & MLP & 0.8635 $\pm$ 0.0032 & 0.8175 $\pm$ 0.0059 & 0.9342 & 0.9494 $\pm$ 0.0046 & \third{0.9565 $\pm$ 0.0060} & 0.9142 $\pm$ 0.0122 \\
& GCN & 0.8031 $\pm$ 0.0425 & 0.8419 $\pm$ 0.0050 & 0.9342 & 0.9494 $\pm$ 0.0046 & 0.9431 $\pm$ 0.0043 & 0.8960 $\pm$ 0.0042 \\
& SAGE & 0.8881 $\pm$ 0.0002 & 0.8372 $\pm$ 0.0082 & 0.9342 & 0.9494 $\pm$ 0.0046 & \first{0.9618 $\pm$ 0.0053} & 0.9087 $\pm$ 0.0064 \\
& RevGAT & 0.8850 $\pm$ 0.0005 & 0.8502 $\pm$ 0.0048 & 0.9342 & 0.9494 $\pm$ 0.0046 & \second{0.9604 $\pm$ 0.0047} & 0.8654 $\pm$ 0.0952 \\
& \textbf{Ensemble} & - & - & - & - & - & 0.9265 $\pm$ 0.0068 \\
\midrule
\textbf{ogbn-arxiv} & MLP & 0.5336 $\pm$ 0.0038 & 0.7308 $\pm$ 0.0006 & 0.7350 & 0.7361 $\pm$ 0.0004 & 0.7587 $\pm$ 0.0015 & 0.7517 $\pm$ 0.0011 \\
& GCN & 0.7182 $\pm$ 0.0027 & 0.7329 $\pm$ 0.0010 & 0.7350 & 0.7361 $\pm$ 0.0004 & 0.7520 $\pm$ 0.0005 & 0.7377 $\pm$ 0.0010 \\
& SAGE & 0.7171 $\pm$ 0.0017 & 0.7435 $\pm$ 0.0014 & 0.7350 & 0.7361 $\pm$ 0.0004 & \third{0.7672 $\pm$ 0.0007} & 0.7596 $\pm$ 0.0040 \\
& RevGAT & 0.7083 $\pm$ 0.0017 & 0.7590 $\pm$ 0.0019 & 0.7350 & 0.7361 $\pm$ 0.0004 & \second{0.7750 $\pm$ 0.0012} & 0.7638 $\pm$ 0.0054 \\
& \textbf{Ensemble} & - & - & - & - & - & \first{0.7777 $\pm$ 0.0019} \\
\midrule
\textbf{ogbn-products} & MLP & 0.5385 $\pm$ 0.0017 & 0.6125 $\pm$ 0.0078 & 0.7440 & 0.7297 $\pm$ 0.0023 & 0.7878 $\pm$ 0.0082 & 0.7277 $\pm$ 0.0054 \\
& GCN & 0.7052 $\pm$ 0.0051 & 0.6977 $\pm$ 0.0042 & 0.7440 & 0.7297 $\pm$ 0.0023 & 0.7996 $\pm$ 0.0041 & 0.7679 $\pm$ 0.0109 \\
& SAGE & 0.6913 $\pm$ 0.0026 & 0.6869 $\pm$ 0.0011 & 0.7440 & 0.7297 $\pm$ 0.0023 & \third{0.8137 $\pm$ 0.0043} & 0.7795 $\pm$ 0.0012 \\
& RevGAT & 0.6964 $\pm$ 0.0017 & 0.7189 $\pm$ 0.0030 & 0.7440 & 0.7297 $\pm$ 0.0023 & \first{0.8234 $\pm$ 0.0036} & 0.8083 $\pm$ 0.0051 \\
& \textbf{Ensemble} & - & - & - & - & - & \second{0.8140 $\pm$ 0.0033}\\
\midrule
\textbf{tape-arxiv23} & MLP & 0.6202 $\pm$ 0.0064 & 0.5574 $\pm$ 0.0032 & 0.7356 & 0.7358 $\pm$ 0.0006 & \third{0.8385 $\pm$ 0.0246} & 0.7940 $\pm$ 0.0022 \\
& GCN & 0.6341 $\pm$ 0.0062 & 0.5672 $\pm$ 0.0061 & 0.7356 & 0.7358 $\pm$ 0.0006 & 0.8080 $\pm$ 0.0215 & 0.7678 $\pm$ 0.0024 \\
& SAGE & 0.6430 $\pm$ 0.0037 & 0.5665 $\pm$ 0.0032 & 0.7356 & 0.7358 $\pm$ 0.0006 & \second{0.8388 $\pm$ 0.0264} & 0.7894 $\pm$ 0.0024 \\
& RevGAT & 0.6563 $\pm$ 0.0062 & 0.5834 $\pm$ 0.0038 & 0.7356 & 0.7358 $\pm$ 0.0006 & \first{0.8423 $\pm$ 0.0256} & 0.7880 $\pm$ 0.0023 \\
& \textbf{Ensemble} & - & - & - & - & - & 0.8029 $\pm$ 0.0020 \\
\bottomrule
\end{tabular}
\caption{Node classification accuracy for the Cora, PubMed, ogbn-arxiv, ogbn-products, and tape-arxiv23 datasets. The experiment is run over four seeds, with mean accuracy and standard deviation shown. The best results are coloured green (first), yellow (second), and orange (third). For $h_{\text{STAGE}}$, we use SFR-Embedding-Mistral as the embedding model on TA features only, and the simple task instruction to bias the embeddings. We adapt the table from \cite{he2024harnessing} and include our results.}
\label{tab:main-results}
\end{table*}

We investigate the performance of STAGE over five TAG benchmarks: \textit{ogbn-arxiv} \cite{hu2021open}, a dataset of arXiv papers linked by citations; \textit{ogbn-products} \cite{hu2021open}, representing an Amazon product co-purchasing network; \textit{PubMed} \cite{Sen_Namata_Bilgic_Getoor_Galligher_Eliassi-Rad_2008}, a citation network of diabetes-related scientific publications; \textit{Cora} \cite{McCallum2000AutomatingTC}, a dataset of scientific publications categorized into one of seven classes; and \textit{tape-arxiv23} \cite{he2024harnessing}, focusing on arXiv papers published after the 2023 knowledge cut-off for GPT3.5. We use the subset of \textit{ogbn-products} provided by \cite{he2024harnessing}. Further details can be found in appendix Table \ref{tab:dataset-stats}.

For each experiment using Cora, PubMed or \textit{tape-arxiv23}, 60\% of the data was allocated for training, 20\% for validation, and 20\% for testing. For the \texttt{ogbn-arxiv} and \texttt{ogbn-products} datasets, we adopted the standard train/validation/test split provided by the Open Graph Benchmark (OGB)\footnote{\url{https://ogb.stanford.edu/}} \cite{hu2021open}.

Our main results can be seen in Table \ref{tab:main-results}. Multiple GNN models are trained using embeddings from a pre-trained LLM as node features. We ensemble the predictions across model architectures by taking the mean prediction. 

\begin{table*}[h!]
\centering
\setlength{\tabcolsep}{4pt}
\small
\begin{tabular}{lcccccc}
\toprule
\textbf{Dataset} & \textbf{Method} & $h_{\text{no instruction}}$ & $h_{\text{task instruction}}$ & $h_{\text{graph-aware-instruction}}$ \\
\midrule
\textbf{Cora} & MLP & 0.7772 $\pm$ 0.0205 & 0.7680 $\pm$ 0.0228 & 0.7763 $\pm$ 0.0193 \\
& GCN & 0.8612 $\pm$ 0.0121 & 0.8704 $\pm$ 0.0105 & 0.8718 $\pm$ 0.0085 \\
& SAGE & \cellcolor{orange!40}{0.8833 $\pm$ 0.0125} & 0.8722 $\pm$ 0.0063 & 0.8704 $\pm$ 0.0109 \\
& RevGAT & 0.8630 $\pm$ 0.0119 & 0.8639 $\pm$ 0.0129 & 0.8676 $\pm$ 0.0125 \\
& \textbf{Ensemble} & \cellcolor{green!40}{0.8930 $\pm$ 0.0086} & 0.8824 $\pm$ 0.0155 & \cellcolor{yellow!40}{0.8875 $\pm$ 0.0118} \\
\midrule
\textbf{PubMed} & MLP & \cellcolor{orange!40}{0.9305 $\pm$ 0.0052} & 0.9142 $\pm$ 0.0122 & 0.9185 $\pm$ 0.0145 \\
& GCN & 0.9021 $\pm$ 0.0034 & 0.8960 $\pm$ 0.0042 & 0.8978 $\pm$ 0.0046 \\
& SAGE & 0.9268 $\pm$ 0.0052 & 0.9087 $\pm$ 0.0064 & 0.9126 $\pm$ 0.0024 \\
& RevGAT & 0.8637 $\pm$ 0.0942 & 0.8654 $\pm$ 0.0952 & 0.9211 $\pm$ 0.0022 \\
& \textbf{Ensemble} & \cellcolor{green!40}{0.9358 $\pm$ 0.0035} & 0.9265 $\pm$ 0.0068 & \cellcolor{yellow!40}{0.9313 $\pm$ 0.0025} \\
\midrule
\textbf{ogbn-arxiv} & MLP & 0.7417 $\pm$ 0.0015 & 0.7517 $\pm$ 0.0011 & 0.7519 $\pm$ 0.0028 \\
& GCN & 0.7336 $\pm$ 0.0029 & 0.7377 $\pm$ 0.0010 & 0.7367 $\pm$ 0.0045 \\
& SAGE & 0.7515 $\pm$ 0.0027 & 0.7596 $\pm$ 0.0040 & 0.7559 $\pm$ 0.0039 \\
& RevGAT & 0.7629 $\pm$ 0.0035 & 0.7638 $\pm$ 0.0054 & 0.7607 $\pm$ 0.0011 \\
& \textbf{Ensemble} & \cellcolor{yellow!40}{0.7745 $\pm$ 0.0013} & \cellcolor{green!40}{0.7777 $\pm$ 0.0019} & \cellcolor{orange!40}{0.7740 $\pm$ 0.0019} \\
\midrule
\textbf{ogbn-products} & MLP & 0.6841 $\pm$ 0.0054 & 0.7277 $\pm$ 0.0054 & 0.7163 $\pm$ 0.0172 \\
& GCN & 0.7367 $\pm$ 0.0068 & 0.7679 $\pm$ 0.0109 & 0.7729 $\pm$ 0.0033 \\
& SAGE & 0.7543 $\pm$ 0.0065 & 0.7795 $\pm$ 0.0012 & 0.7811 $\pm$ 0.0049 \\
& RevGAT & 0.8016 $\pm$ 0.0078 & \third{0.8083 $\pm$ 0.0051} & 0.8000 $\pm$ 0.0078 \\
& \textbf{Ensemble} & 0.7991 $\pm$ 0.0034 & \cellcolor{green!40}{0.8140 $\pm$ 0.0033} & \cellcolor{yellow!40}{0.8090 $\pm$ 0.0037} \\
\midrule
\textbf{tape-arxiv23} & MLP & 0.7803 $\pm$ 0.0014 & 0.7940 $\pm$ 0.0022 & 0.7948 $\pm$ 0.0025 \\
& GCN & 0.7518 $\pm$ 0.0044 & 0.7678 $\pm$ 0.0024 & 0.7703 $\pm$ 0.0025 \\
& SAGE & 0.7702 $\pm$ 0.0022 & 0.7894 $\pm$ 0.0024 & 0.7917 $\pm$ 0.0021 \\
& RevGAT & 0.7880 $\pm$ 0.0047 & 0.7880 $\pm$ 0.0023 & 0.7906 $\pm$ 0.0034 \\
& \textbf{Ensemble} & \third{0.8013 $\pm$ 0.0017} & \second{0.8029 $\pm$ 0.0020} & \first{0.8054 $\pm$ 0.0025} \\
\bottomrule
\end{tabular}
\caption{Node classification accuracy for the Cora, PubMed, ogbn-arxiv, ogbn-products, and tape-arxiv23 datasets, demonstrating the effect of varying an instruction to bias the embeddings from the pre-trained LLM. The experiment is run over four seeds, with mean accuracy and standard deviation shown. The best results are coloured green (first), yellow (second), and orange (third). For all experiments, we use SFR-Embedding-Mistral as the embedding model on TA features only, and the simple task instruction to bias the embeddings.}
\label{tab:instruction-bias-results}
\end{table*}

Node classification accuracy is provided for various datasets, measured across multiple methods and feature types. Each column represents a specific metric or method:

\begin{itemize}[itemsep=0.1em]
    \item \textbf{h\textsubscript{shallow}}: Performance using shallow features, indicating basic attributes provided as part of each dataset
    \item \textbf{h\textsubscript{GIANT}}: Results obtained by using GIANT features as proposed by \cite{chien2022node}, designed to incorporate graph structural information into LM training
    \item \textbf{GPT3.5}: Accuracy when using zero-shot predictions from GPT-3.5-turbo, demonstrating the utility of state-of-the-art language models in a zero-shot setting
    \item \textbf{LM\textsubscript{finetune}}: Performance metrics reported by \cite{he2024harnessing} after finetuning the DeBERTa \cite{he2021deberta} model on labeled nodes from the graph, showing the benefits of supervised finetuning
    \item \textbf{h\textsubscript{TAPE}}: Shows results for the TAPE features \cite{he2024harnessing}, which includes the original textual attributes of the node, GPT-generated predictions for each node, and GPT-generated explanations of ranked predictions to enrich node features.
    \item \textbf{h\textsubscript{STAGE}}: Reflects the model's performance training with node features generated by a pre-trained LLM.
\end{itemize}

\paragraph{Instruction-biased Embeddings}
Textual attributes for each node are passed to the embedding LLM together with a task description which remains constant for every text, prefixing each input with a task-specific system prompt. We evaluated 3 simple task descriptions:
\begin{enumerate}
    \item A short prompt describing the classification task for the text, as used during the pre-training stage of the LLM.
    \item A description of the types of relationships between texts to form a graph, along with the classification task description. Specific graph structure for each node is not included in the prompt, unlike the proposed method from \cite{fatemi2024talk}.
    \item No task description.
\end{enumerate}

\noindent Our findings are summarized in Table \ref{tab:instruction-bias-results}. Further details of the instructions can be found in appendix Table \ref{tab:task_descriptions}.

\paragraph{Parameter-efficient Finetuning}
In Table \ref{tab:peft-results} we investigate the effect of using parameter-efficient finetuning (PEFT) on the pre-trained LLM, as described in \cite{duan2023simteg}. We also compare this against finetuning both the LLM (using PEFT) and the GNN in unison.

\begin{table*}[h!]
\centering
\setlength{\tabcolsep}{4pt}
\small
\begin{tabular}{lcccc}
\toprule
\textbf{Dataset} & \textbf{LLM + GNN Ensemble} & \textbf{LLM\textsubscript{finetuned}} & \textbf{LLM\textsubscript{finetuned} + GNN Ensemble} \\
\midrule
\textbf{Cora} & 0.8824 $\pm$ 0.0155 & 0.8063 & \textbf{0.8856} \\
\textbf{PubMed} & 0.9265 $\pm$ 0.0068 & 0.9513 & \textbf{0.9559} \\
\textbf{ogbn-arxiv} & 0.7777 $\pm$ 0.0019 & 0.7666 & \textbf{0.7813} \\
\textbf{ogbn-products} & 0.8140 $\pm$ 0.0033 & 0.8020 & \textbf{0.8257} \\
\textbf{tape-arxiv23} & 0.8029 $\pm$ 0.0020 & 0.8021 & \textbf{0.8095} \\
\bottomrule
\end{tabular}
\caption{Effect of using parameter-efficient finetuning (PEFT) on the pre-trained LLM, as described in \cite{duan2023simteg}. Comparison of GNN-only trained, LLM finetuned without GNNs, and LLM and GNN trained separately. The best results are highlighted in bold.}
\label{tab:peft-results}
\end{table*}

\paragraph{Embedding Model Type}
In Table \ref{tab:combined-embedding-model-comparison}, we compare the results when using different pre-trained LLMs as the text encoder.

\begin{table*}[h!]
\centering
\setlength{\tabcolsep}{4pt}
\small
\begin{tabular}{lcccccc}
\toprule
\textbf{Dataset} & \textbf{Method} & \textbf{SFR-Embedding-Mistral} & \textbf{LLM2Vec} & \textbf{gte-Qwen1.5-7B-instruct} \\
\midrule
\textbf{Cora} & MLP & 0.7680 $\pm$ 0.0228 & 0.8026 $\pm$ 0.0141 & 0.7389 $\pm$ 0.0136 \\
& GCN & 0.8704 $\pm$ 0.0105 & 0.8778 $\pm$ 0.0046 & 0.8621 $\pm$ 0.0105 \\
& SAGE & 0.8722 $\pm$ 0.0063 & 0.8773 $\pm$ 0.0062 & 0.8658 $\pm$ 0.0049 \\
& RevGAT & 0.8639 $\pm$ 0.0129 & 0.8810 $\pm$ 0.0033 & 0.8408 $\pm$ 0.0076 \\
& \textbf{Ensemble} & \third{0.8824} $\pm$ 0.0155 & \first{0.8898 $\pm$ 0.0066} & 0.8686 $\pm$ 0.0024 \\
\midrule
& Simple-GCN & 0.7389 $\pm$ 0.0120 & 0.6983 $\pm$ 0.0120 & 0.7491$\pm$ 0.0166 \\
& SIGN & 0.8819 $\pm$ 0.0074 & \second{0.8856 $\pm$ 0.0083} & 0.8575 $\pm$ 0.0157 \\
\midrule
\textbf{PubMed} & MLP & 0.9142 $\pm$ 0.0122 & \second{0.9321 $\pm$ 0.0013} & 0.8808 $\pm$ 0.0107 \\
& GCN & 0.8960 $\pm$ 0.0042 & 0.8996 $\pm$ 0.0011 & 0.8591 $\pm$ 0.0041 \\
& SAGE & 0.9087 $\pm$ 0.0064 & 0.9231 $\pm$ 0.0056 & 0.8733 $\pm$ 0.0051 \\
& RevGAT & 0.8654 $\pm$ 0.0952 & \third{0.9312 $\pm$ 0.0026} & 0.8754 $\pm$ 0.0010 \\
& \textbf{Ensemble} & 0.9265 $\pm$ 0.0068 & \first{0.9357 $\pm$ 0.0031} & 0.8941 $\pm$ 0.0041 \\
\midrule
& Simple-GCN & 0.7505 $\pm$ 0.0048 & 0.7400 $\pm$ 0.0037  & 0.7472 $\pm$ 0.0076 \\
& SIGN & 0.8868 $\pm$ 0.0062 & 0.9004 $\pm$ 0.0038 & 0.8611 $\pm$ 0.0084 \\
\midrule
\textbf{ogbn-arxiv} & MLP & 0.7517 $\pm$ 0.0011 & 0.7331 $\pm$ 0.0033 & 0.7603 $\pm$ 0.0011 \\
& GCN & 0.7377 $\pm$ 0.0010 & 0.7324 $\pm$ 0.0014 & 0.7369 $\pm$ 0.0022 \\
& SAGE & 0.7596 $\pm$ 0.0040 & 0.7428 $\pm$ 0.0039 & 0.7664 $\pm$ 0.0029 \\
& RevGAT & 0.7638 $\pm$ 0.0054 & 0.7529 $\pm$ 0.0044 & \third{0.7738 $\pm$ 0.0009} \\
& \textbf{Ensemble} & \second{0.7777 $\pm$ 0.0019} & 0.7701 $\pm$ 0.0018 & \first{0.7817 $\pm$ 0.0011} \\
\midrule
& Simple-GCN & 0.3337 $\pm$ 0.0107 & 0.3614 $\pm$ 0.0039 & 0.3463 $\pm$ 0.0181 \\
& SIGN & 0.6150 $\pm$ 0.0182 & 0.6035  $\pm$ 0.0084  & 0.6285 $\pm$ 0.0114 \\
\midrule
\textbf{ogbn-products} & MLP & 0.7277 $\pm$ 0.0054 & 0.6913 $\pm$ 0.0052 & 0.7231 $\pm$ 0.0050 \\
& GCN & 0.7679 $\pm$ 0.0109 & 0.7479 $\pm$ 0.0128 & 0.7701 $\pm$ 0.0117 \\
& SAGE & 0.7795 $\pm$ 0.0012 & 0.7496 $\pm$ 0.0163 & 0.7921 $\pm$ 0.0069 \\
& RevGAT & \third{0.8083 $\pm$ 0.0051} & 0.7883 $\pm$ 0.0014 & 0.7955 $\pm$ 0.0096 \\
& \textbf{Ensemble} & \first{0.8140 $\pm$ 0.0033} & 0.7908 $\pm$ 0.0045 & \second{0.8104 $\pm$ 0.0041} \\
\midrule
& Simple-GCN & 0.6216 $\pm$ 0.0052 & 0.6040 $\pm$ 0.0039 & 0.6219 $\pm$ 0.0039 \\
& SIGN & 0.6668 $\pm$ 0.0078 & 0.6621 $\pm$ 0.0009 & 0.6698 $\pm$ 0.0010 \\
\midrule
\textbf{tape-arxiv23} & MLP & 0.7940 $\pm$ 0.0022 & 0.7772 $\pm$ 0.0033 & \third{0.8008 $\pm$ 0.0018} \\
& GCN & 0.7678 $\pm$ 0.0024 & 0.7541 $\pm$ 0.0042 & 0.7746 $\pm$ 0.0025 \\
& SAGE & 0.7894 $\pm$ 0.0024 & 0.7677 $\pm$ 0.0018 & 0.7975 $\pm$ 0.0016 \\
& RevGAT & 0.7880 $\pm$ 0.0023 & 0.7840 $\pm$ 0.0058 & 0.7954 $\pm$ 0.0028 \\
& \textbf{Ensemble} & \second{0.8029 $\pm$ 0.0020} & 0.7967 $\pm$ 0.0037 & \first{0.8065 $\pm$ 0.0022} \\
\midrule
& Simple-GCN & 0.2516 $\pm$ 0.0027 & 0.2451 $\pm$ 0.0004 & 0.258 $\pm$ 0.0011 \\
& SIGN & 0.7186 $\pm$ 0.0041 & 0.6804 $\pm$ 0.0041 & 0.733 $\pm$ 0.0009 \\
\bottomrule
\end{tabular}
\caption{Node classification accuracy for the Cora, PubMed, ogbn-arxiv, ogbn-products, and tape-arxiv23 datasets, demonstrating the effect of changing the pre-trained LLM text encoder. The experiment is run over four seeds, with mean accuracy and standard deviation shown. The best results are coloured green (first), yellow (second), and orange (third). For all experiments, we use TA features only, and the simple task instruction to bias the embeddings.}
\label{tab:combined-embedding-model-comparison}
\end{table*}

\paragraph{Diffusion GNNs}
Included in Table \ref{tab:combined-embedding-model-comparison}, we study the performance of using SimpleGCN and SIGN models individually.
Model selection and implementation details can be found in the appendix sections \ref{sec:implementing-diffusion-operators} and \ref{sec:model-selection}.

\paragraph{Ablation Study}
To study the impact of each component in the GNN ensemble, we perform a detailed ablation study. The results can be found in \ref{tab:ablation}.

\section{Analysis}
\label{sec:analysis}

\paragraph{Main Results (Table \ref{tab:main-results})}

We find that ensembling GNNs always leads to superior performance across datasets when taking the STAGE approach.

Despite the reduced computational resources and training data requirements, the STAGE method remains highly competitive across all benchmarks. The ensemble STAGE approach lags behind the TAPE pipeline by roughly 5\% on Cora, 3.5\% on Pubmed, 0.8\% on ogbn-products, and 4\% on tape-arxiv23. This is a strong result when we consider that STAGE involves training only the GNN ensemble, whereas TAPE also requires two finetuned LMs to generate node features. We see marginally superior results on the ogbn-arxiv dataset using the ensemble STAGE approach.


\paragraph{Instruction-biased Embedding Results (Table \ref{tab:instruction-bias-results})}

From our findings we conclude that varying the instructions to bias embeddings has little effect on downstream node classification performance for the models we evaluated. We note that while the authors of all embedding models recommend providing instructions along with input text in order to avoid degrading performance, we did not measure a performance improvement in our experiments.

This experiment further supports our claim that an ensemble approach improves robustness across datasets and methods of node feature generation.

\paragraph{PEFT Results (Table \ref{tab:peft-results})}

Finetuning each LLM gave marginal performance improvements across all datasets to varying degrees; we see the largest improvement on pubmed (3\%). It is of note that finetuning significantly increases the number of trainable parameters (see Table \ref{tab:model_params}) and total training time. Specifically, PEFT for 7B embedding models has over 20 million trainable parameters. On a single A100 GPU, training runs lasted 6 hours on \texttt{ogbn-arxiv}.

\paragraph{LLM Embedding Model Comparison (Table \ref{tab:combined-embedding-model-comparison})}

All three LLM embedding models demonstrated comparable performance on the graph tasks, with each model exhibiting marginally better results on different datasets. Notably, there was no clear winner among them. The LLM2Vec model exhibited slightly weaker performance on the larger datasets (ogbn-arxiv, ogbn-products, tape-arxiv23), while it was marginally stronger on the smaller datasets (Cora, PubMed).

Ensembling the GNN models consistently ranked among the top three models across all three LLM embedding models, delivering an average performance increase of 1\%. Among the individual GNN architectures, RevGAT consistently demonstrated superior performance.

\paragraph{Diffusion-pattern GNN Results (Table \ref{tab:combined-embedding-model-comparison})}The diffusion-based GNNs yielded variable results across datasets. Specifically, SIGN emerged as the second-best performer on the Cora dataset. As expected, SIGN consistently outperformed Simple-GCN, given that it generalizes the latter. 
Due to its low training time, SIGN is a viable candidate for large datasets, although careful tuning of its hyper-parameters is recommended for optimal performance.

\paragraph{Ablation Study Results (Table \ref{tab:ablation})}

From our ablation study we observe that no individual GNN model outperforms any ensemble of models on any dataset. Additionally, we find that the full ensemble of MLP, GCN, SAGE and RevGAT achieve the highest and most stable accuracy scores across datasets.

\paragraph{Scalability}

An important advantage of STAGE is the lack of finetuning necessary to achieve strong results. This lies in contrast to approaches such as TAPE \cite{he2024harnessing} and SimTeG \cite{duan2023simteg}, both of which require finetuning at least one LM. Training an ensemble of GNNs and MLP head over the ogbn-arxiv dataset can be performed on a single consumer-grade GPU in less than 5 minutes. This is illustrated in Figure \ref{fig:tradeoff} where we compare the relationship between training time and accuracy for a number of SoTA node classification approaches. When using SIGN diffusion, training time was under 12 seconds for the ogbn-arxiv, but this came at a performance cost.
Moreover, TAPE relies on text-level enhancement via LLM API calls, which adds a new dimension of cost and rate-limiting\footnote{\url{https://platform.openai.com/docs/guides/rate-limits}} to consider when adapting to other datasets.

\section{Conclusions}
\label{sec:conclusions}

This work introduces STAGE, a method to use pre-trained LLMs as text encoders in TAG tasks without the need for finetuning, significantly reducing computational resources and training time. Additional gains can be achieved through parameter-efficient finetuning of the LLM. Data augmentation, which is orthogonal to our approach, could improve performance with general-purpose text embedding models. However, it likely remains intractable for many large-scale datasets due to the need to query a large model for each node.

We also demonstrate the effect of diffusion operators \cite{frasca2020sign} on node classification performance, decreasing TAG pipeline training time substantially. We aim to examine the scalability of diffusion-pattern GNNs on larger datasets in later work.

Future work may aim to refine the integration of LLM encoders with GNN heads. Potential strategies include an Expectation-Maximization approach or a joint model configuration \cite{glem}. A significant challenge is the requirement for large, variable batch sizes during LLM finetuning due to current neighborhood sampling techniques, which necessitates increased computational power. We anticipate that overcoming these limitations will make future research more accessible and expedite iterations.


\bibliography{custom}


\appendix

\section{Appendix}
\label{sec:appendix}

\section{Negative Results}

\paragraph{Co-training LLM and GNN:} In a similar approach to iterative methods, we investigated co-training the LLM and GNN on the \texttt{ogbn-arxiv} node classification task to facilitate a shared representation space. This proved unfeasible due to the memory requirements exceeding the capacity of one A100 GPU.



\section{Implementation of Diffusion Operators}
\label{sec:implementing-diffusion-operators}
We implement diffusion operators from two methods, Simple-GCN \cite{simpleGCNwu} and SIGN \cite{frasca2020sign}. In the case of SIGN, the authors omit implementation details of the operators, so we include them here.

Let $A$ denote the adjacency matrix of a possibly directed graph $G$, $X$ its node features, and $D$ the diagonal degree matrix of $G$.

We denote the \textit{random-walk normalized} adjacency $A_{\text{RW}} \&\coloneqq AD^{-1}$ and the \textit{GCN-normalized} adjacency  \cite{kipf2017semisupervised} 
\begin{align}
    A_{\text{GCN}} &\coloneqq \left(D + I\right)^{-1/2}\left(A + I\right)\left(D + I\right)^{-1/2}
\end{align}

The \textit{Personalized PageRank} matrix is then given by \cite{gasteiger2022predict}:
\begin{align}
    A_{\text{PPR}} &\coloneqq \alpha\left(I_n - \left(1 - \alpha\right)A_{\text{RW}}\right)^{-1} \label{eq:ppr}
\end{align}

And we denote the \textit{triangle-based} adjacency matrix by $\mathbf{A}_{\Delta}$, where $\left(A_{\Delta}\right)_{ij}$ counts the number of directed triangles in $G$ that contain the edge $(i, j)$

Diffusion is applied to node features $X$ by matrix multiplication. Simple-GCN takes a power $k$ of $A_{\text{GCN}}$ as its diffusion operator, whilst SIGN diffusion generalizes this to concatenate powers of $A_{\text{GCN}}$, $A_{\text{PPR}}$ and $A_{\Delta}$.

Diffusion can be calculated efficiently if sparse-matrix-sparse-matrix multiplication is avoided. For both SIGN and Simple-GCN, the order of operations for applying a power of an operator $A_{\text{op}}$ should be
\begin{align}
    \underbrace{A_{\text{op}}(A_{\text{op}}(...(A_{\text{op}}(X))...)}_{k\;\text{times}} \label{eq:recursive-diff}
\end{align}
as opposed to $(A_{\text{op}}^k) X$, where the operator matrix $A_{op}$ is feasible to calculate, since the former avoids sparse matrix multiplication. In SIGN, the recursive nature of eq.\ref{eq:recursive-diff} can be exploited to reuse results for calculating successive powers.

In the case of personalized pagerank diffusion,  we first use a trick from \cite{gasteiger2022predict} to approximate the diffused features of personalized pagerank matrix $A_{\text{PPR}}X$ in linear time and avoid calculative $A_{\text{PPR}}$ directly, by viewing eq.\ref{eq:ppr} as \textit{topic-sensitive} PageRank \cite{topicSensitivePR}. We use the random-walk normalized adjacency matrix.

The following power iteration approximates $A_{\text{PPR}}X$ (notation from \cite{gasteiger2022predict}):
\begin{align*}
    Z^{(0)} &\coloneqq X \\
    Z^{(k+1)} &\coloneqq (1-\alpha)AZ^{(k)} + \alpha X
\end{align*}
    
To compute the $n$th diffused power, we repeat the process $n$ times:
\begin{align*}
Z^{(0)}_0 &= X \\
Z^{(0)}_{n+1} &= \lim\limits_{k \to \inf} Z^{(k)}_n \\
\end{align*}
Lastly, for triangle-based diffusion, we count triangles using linear algebra. For unweighted $A$ we perform a single sparse matrix multiplication to obtain $A^2$, in which element $(i, j)$ counts the directed paths in $G$ for node $i$ to node $j$. We then calculate
\begin{align*}
    A_{\Delta} = A^T \odot A^2
\end{align*}
where $\odot$ denotes the Hadamard product, which can be efficiently calculated for sparse matrices. We then normalize and diffuse features over powers of $A_{\Delta}$ in the same fashion as for $A_{GCN}$.

An implementation of these operators as GraphBLAS \cite{GraphBLAS7} code is published alongside this paper.

\subsection{Parallelism of diffusion operators}
All operations above can be be parallelized across columns of $X$, either keeping $A$ in shared memory on one machine or keeping a copy on each executor in a distributed computing infrastructure like Apache Spark.

\section{Preprocessing \& Model Selection for Diffusion Operators}
\label{sec:model-selection}
For Simple-GCN \cite{simpleGCNwu}, we set the degree $k$ by selecting the highest validation accuracy from $k=2,3,4$, of which $k=2$ had the highest accuracy in each case. For SIGN \cite{frasca2020sign}, we choose $s$, $p$, $t$ from the highest validation accuracy amongst $(3,0,0)$ $(3,0,1)$ $(3,3,0)$, $(4,2,1)$ $(5,3,0)$. For \textbf{Cora} and \textbf{PubMed}, $(4,2,1)$ was chosen, and for \textbf{ogbn-arxiv}, \textbf{ogbn-products}, and \textbf{tape-arxiv23} $(3,3,0)$ was chosen. We chose the number of layers for the Inception NLP to match the number of layers in other GNNs tested, 4. We did not perform additional hyper-parameter tuning. When preprocessing the embeddings, we centered and scaled the data to unit variance for Simple-GCN and SIGN only.

\section{Model Trainable Parameters}

\begin{table}[ht!]
    \centering
    \begin{tabular}{lcccc}
        \toprule
        \textbf{Model} & \textbf{Trainable Parameter Count} \\
        \midrule
        RevGAT & 3,457,678 \\
        GCN & 559,111 \\
        SAGE & 1,117,063 \\
        MLP & 117,767 \\
        Simple-GCN & 24,111 \\
        SIGN-(3,3,0) & 500,271 \\
        SIGN-(4,2,1) & 582,575 \\
        PEFT 7B LLM & >20M \\
        \bottomrule
    \end{tabular}
    \caption{Trainable parameter counts for different models. 7B LLM refers to all finetuned LLM embedding models used during experiments (see Section \ref{sec:text-embedding-retrieval})}
    \label{tab:model_params}
\end{table}

\newpage

\section{Ablation Study}

To study the effect each model has on the GNN ensemble step of STAGE, we perform a detailed ablation study. The results are shown in Table \ref{tab:ablation}.

\begin{table*}[htbp!]
\centering
\small
\begin{tabular}{lcccccc}
\toprule
\textbf{Method} & \textbf{Cora} & \textbf{PubMed} & \textbf{ogbn-arxiv} & \textbf{ogbn-products} & \textbf{tape-arxiv23} \\
\midrule
Full Ensemble & \second{0.8824 $\pm$ 0.0155} & \second{0.9265 $\pm$ 0.0068} & \first{0.7777 $\pm$ 0.0019} & \first{0.8140 $\pm$ 0.0033} & \first{0.8029 $\pm$ 0.0020} \\
No MLP & \first{0.8838 $\pm$ 0.0039} & 0.9239 $\pm$ 0.0036 & \second{0.7748 $\pm$ 0.0012} & 0.8093 $\pm$ 0.0021 & 0.8015 $\pm$ 0.0010 \\
No GCN & 0.8685 $\pm$ 0.0209 & 0.9240 $\pm$ 0.0076 & 0.7731 $\pm$ 0.0017 & \third{0.8100 $\pm$ 0.0038} & \second{0.8028 $\pm$ 0.0023} \\
No SAGE & 0.8759 $\pm$ 0.0207 & \third{0.9258 $\pm$ 0.0110} & \third{0.7739 $\pm$ 0.0020} & \second{0.8116 $\pm$ 0.0045} & \third{0.8021 $\pm$ 0.0035} \\
No RevGAT & \third{0.8764 $\pm$ 0.0180} & \first{0.9272 $\pm$ 0.0052} & 0.7717 $\pm$ 0.0007 & 0.8029 $\pm$ 0.0036 & 0.7985 $\pm$ 0.0018 \\
\midrule
Best Individual & 0.8722 $\pm$ 0.0063 & 0.9142 $\pm$ 0.0122 & 0.7638 $\pm$ 0.0054 & 0.8083 $\pm$ 0.0051 & 0.7880 $\pm$ 0.0023 \\
\midrule
\textbf{Best Individual Model} & \textbf{SAGE} & \textbf{MLP} & \textbf{RevGAT} & \textbf{RevGAT} & \textbf{RevGAT} \\
\bottomrule
\end{tabular}
\caption{Ablation study results for the ensemble model on various datasets. The table shows the accuracy when each component is removed from the ensemble. The experiment is run over four seeds, with mean accuracy and standard deviation shown. The best results are coloured green (first), yellow (second), and orange (third). For all experiments, we use SFR-Embedding-Mistral as the embedding model on TA features only, and the simple task instruction to bias the embeddings.}
\label{tab:ablation}
\end{table*}

\section{Datasets}

In this section, we describe the characteristics of the node classification datasets we used during our work. The statistics are shown in Table \ref{tab:dataset-stats}.

\begin{table*}[htbp!]
\centering
\begin{tabular}{lcccccc}
\toprule
\textbf{Dataset} & \textbf{Node Count} & \textbf{Edge Count} & \textbf{Task} & \textbf{Metric} \\
\midrule
Cora \cite{McCallum2000AutomatingTC} & 2,708 & 5,429 & 7-class classif. & Accuracy \\
Pubmed \cite{Sen_Namata_Bilgic_Getoor_Galligher_Eliassi-Rad_2008} & 19,717 & 44,338 & 3-class classif. & Accuracy \\
ogbn-arxiv \cite{hu2021open} & 169,343 & 1,166,243 & 40-class classif. & Accuracy \\
ogbn-products \cite{hu2021open} (subset) & 54,025 & 74,420 & 47-class classif. & Accuracy \\
tape-arxiv23 \cite{he2024harnessing} & 46,198 & 78,548 & 40-class classif. & Accuracy & \\
\bottomrule
\end{tabular}
\caption{Statistics of the TAG datasets}
\label{tab:dataset-stats}
\end{table*}

\section{Instruction-biased Embeddings}

In Table \ref{tab:task_descriptions} we list the specific instructions used to 655
investigate the effect of biasing embeddings.

\begin{table*}[h!]
\centering
\begin{tabular}{|l|l|p{8cm}|}
\hline
\textbf{Dataset} & \textbf{Prompt Type} & \textbf{Prompt} \\ \hline
ogbn-arxiv, arxiv\_2023, cora, pubmed & Simple Task & Identify the main and secondary category of Arxiv papers based on the titles and abstracts. \\ \hline
ogbn-arxiv, arxiv\_2023, cora, pubmed & Graph-Aware & Identify the main and secondary category of Arxiv papers based on the titles and abstracts. Your predictions will be used in a downstream graph-based prediction that for each paper can learn from your predictions of neighboring papers in a graph as well as the predictions for the paper in question. Papers in the graph are connected if one cites the other. \\ \hline
ogbn-products & Simple Task & Identify the main and secondary category of this product based on the titles and description. \\ \hline
ogbn-products & Graph-Aware & Identify the main and secondary category of this product based on the titles and description. Your predictions will be used in a downstream graph-based prediction that for each product can learn from your predictions of neighboring products in a graph as well as the predictions for the paper in question. Products in the graph are connected if they are purchased together. \\ \hline
\end{tabular}
\caption{Task descriptions for embedding bias across various datasets.}
\label{tab:task_descriptions}
\end{table*}


\end{document}